\documentclass[letterpaper, 10 pt, conference]{ieeeconf}  

\IEEEoverridecommandlockouts                              

\usepackage{amsmath, amssymb, graphics, graphicx, times, epsfig, subcaption, color, algorithm2e, array}

\title{\LARGE  FPR -- Fast Path Risk Algorithm to Evaluate Collision Probability}

\author{A. Blake$^{*1}$, A. Bordallo$^{*}$, K. Brestnichki$^{*}$, M. Hawasly$^{*}$, S. Penkov$^{*}$, S. Ramamoorthy$^{*}$, A. Silva$^{*}$%
\thanks{$^{*}$Five.AI, 20 Cambridge Pl, Cambridge CB2 1NS, UK.
        {\tt\small \{ab,alex,kamen,majd,svet,ram,alex.silva\}@five.ai}}%
  \thanks{$^{1}$Corresponding author.}
  }

\begin{document}

\maketitle

\begin{abstract}
As mobile robots and autonomous vehicles become increasingly  prevalent in human-centred environments, there is a need to control the risk of collision. Perceptual modules, for example machine vision, provide  {\em uncertain} estimates of object location. In that context, the frequently made assumption of an exactly known free-space  is invalid. Clearly, no paths can be {\em guaranteed} to be collision free.  Instead, it is necessary to compute the probabilistic {\em  risk of collision} on any proposed path.

The FPR algorithm, proposed here, efficiently calculates an upper bound on the risk of collision for a robot moving on the plane. That computation orders candidate trajectories according to (the bound on) their degree of risk. Then paths within a user-defined threshold of primary risk could be selected according to secondary criteria such as comfort and efficiency.

The key contribution of this paper is the FPR algorithm and its `convolution trick' to factor the integrals used to bound the risk of collision. As a consequence of the convolution trick, given $K$ obstacles and $N$ candidate paths, the computational load is reduced from the naive $O(NK)$, to the qualitatively faster $O(N+K)$.
\end{abstract}

\section{INTRODUCTION}

Mobile robotic systems that move autonomously in complex environments are becoming more prevalent. However, the perceptual input available to a mobile robot, for example from computer vision, is uncertain. Therefore it is {\em not possible to certify that a path is collision-free}. When such robots perform complex manoeuvres among obstacles, absolute safety  therefore cannot be guaranteed~\cite{r19}. But instead, robots can operate within a controlled level of risk of collision~\cite{r1}.

Typically, the environment is perceived through sensors such as stereo vision or LIDAR. Uncertainty  arises  directly from sensor noise, and then indirectly through perception  algorithms that detect discrete
obstacles~\cite{r22, r23} or impassable terrain. It is therefore realistic to expect each perception module to output a {\em probability distribution}~\cite{r2} over pose, for each detected object. Then, given these random variables from detector outputs, candidate paths can be assessed for the risk of collision.
\begin{figure}[htbp]
\centering
\vspace{3mm}\includegraphics[width=0.45\textwidth]{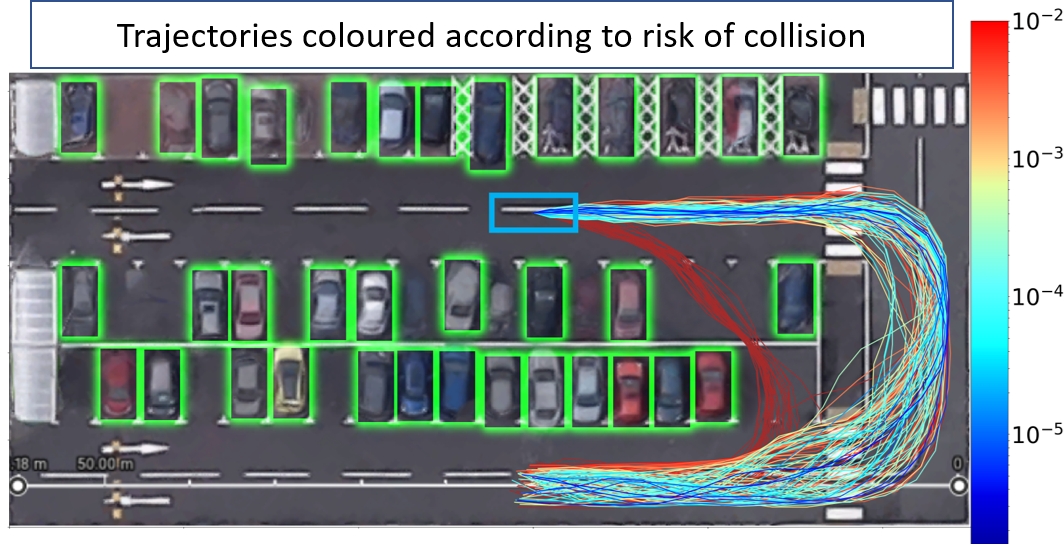}
\caption{\small{\bf Robot motion problem.} Data taken from an aerial view of a car-park. The obstacles here are 35 cars with shapes given by the green bounding boxes, and uncertainties in location visualised as a green halo. The robot vehicle (blue) has a set of candidate paths, evaluated here as high risk (red) to low risk (blue) according to the scale shown. Bounds on collision risk are represented by different colours as on the logarithmic scale shown. Some higher risk paths involve squeezing through a narrow gap.} \label{f:trajectory-risk}
\end{figure}
The FPR algorithm introduced here efficiently computes a bound $F_{\rm D}$ on risk of collision. Note that a bound on risk would not be useful for determining an optimal path. However the problem in this paper is different: to select paths whose risk fall below a certain threshold. For that a bound is entirely usable.

\subsection{Specifying the Problem}
It is assumed that a static freespace in the plane is defined deterministically, and that a robot of known shape $A$ translates and rotates in that freespace. In addition, discrete ``tethered'' obstacles $k=1\ldots,K$ of known shape and uncertain location are perceived by the robot's sensor systems. Of course there is a great deal of work in Robotics addressing uncertain estimation of robot location. This has been so successful (e.g.\cite{Cremers13}) that here we assume that uncertainty in robot location is negligible compared with uncertainty in the locations of perceived obstacles.

The problem is then, over a (short) time-interval $t\in\,[0,\ldots,T]$ to:
\begin{enumerate}
\item{\bf Generate} $N$ candidate paths in configuration space $\Re^2 \times S$ for the robot. Each such path then sweeps out a shape $A$ in the plane $\Re^2$.
\item{\bf Bound} the risk of collision: a bound $F_{\rm D}$ on risk is computed for each candidate path.
\end{enumerate}
The main contribution of the paper relates not to 1. above, for which  off-the-shelf methods are used, but to 2. where we introduce a novel, fast computation of a bound on the risk of collision for a given path. Its computational complexity is $O(N+K)$, compared with the naive $O(NK)$. Once the bound $F_{\rm D}$ has been computed for the first path, the cost of computing the bound for subsequent paths is {\em independent} of the {\em number of obstacles} $K$.

The following inputs to the risk computation are assumed:
\begin{enumerate}
\item{\bf Freespace}: it is assumed that freespace $F$ is initially defined as a subset of $\Re^2$. Then, within $F$, further, tethered obstacles are defined stochastically as below.
\item{\bf Robot}: assumed to have a deterministic spatial extent and to be manoeuvrable in translation and rotation.  Over the interval $t\in\,[0,\ldots,T]$, it sweeps out the set $A$ in the plane.
\item{\bf Obstacle shape}: the $k^{\rm th}$ obstacle is assumed to have deterministic shape  and mean orientation represented by the set $B_k$.
\item{\bf Tethered obstacle}: tethering here means specifying a probability distribution $p_k({\bf r})$ for the location of the $k^{\rm th}$ obstacle, where ${\bf r}=(x, y)$ are coordinates in the plane. This takes into account: i) variability arising from any possible motion over the time-interval $[0,\ldots,T]$; and ii) modelled uncertainty in object detector output. Obstacle locations are assumed to be mutually independent --- arising from  independent sensor observation/detection.
\item{\bf Obstacle rotation:} treated as part of its shape so, to the extent that an obstacle may rotate over the time interval $[0,\ldots,T]$, that rotation must be absorbed in a deterministic expansion of the set $B_k$.
\end{enumerate}
This treatment of obstacle rotation is a limitation of our framework which, however, is reasonable if the time interval $[0,\ldots,T]$ is short, so rotation is limited. For longer time intervals it may be necessary to consider the full $xyt$-space, rather than more simply the $xy$-plane as in this paper. However our treatment is fully general in the rotation of the robot.

\subsection{Existing approaches}

We review some prominent approaches to computing the risk of collision under uncertainty. There are a number of recent approaches to estimating the risk of collision under uncertain robot dynamics \cite{ud1,ud2,ud3}. In our problem it is the environment that is uncertain rather than the dynamics. One approach to this problem involves  casting ``shadows'' around obstacles~\cite{r13} but that does not facilitate the resolution of uncertainty from multiple different sources. Probability density functions can usefully model robot and obstacle uncertainty, as in~\cite{ab2}, which however requires Monte-Carlo computation and has $O(NK)$ complexity.  Bevilacqua et al.~\cite{ab3} model obstacles stochastically, but deal with just one obstacle, and do not allow for sensor or perceptual uncertainty. Empirical probability distributions can also be useful~\cite{ab4} in the case of a single obstacle. Alternatively Althoff et al.~\cite{ab5} elegantly avoid Monte Carlo computation by compiling  stochastic reachability of moving obstacles down to finite Markov Chains, but the risk computation remains $O(NK)$.

Probabilistic Occupancy Grids are an established mechanism for dealing with spatial uncertainty probabilistically~\cite{r3,r10a} and can be used to find paths. However, to calculate the risk of collision along a path with numerous obstacles, a single grid is not enough. Laugier and collaborators~\cite{r10b,r10} show that certain ``Laugier integrals'' (our term) over multiple grids, one grid per point-obstacle, can be combined nonlinearly to compute the total risk of collision. We build on this approach.

An important question is then whether the combination of the Laugier integrals can be simplified somehow, despite the nonlinearity. For example, if the set of obstacles could somehow be replaced by the union of obstacles, that could live on a single grid, it would simplify computation. However, given that obstacles are each defined here not just by their shape but also by the uncertainty in their location, it turns out that constructing a composite obstacle as a trivial union of obstacle shapes is not valid. Therefore, in the FPR algorithm, we derive and justify a non-trivial combination of shape properties and location distributions, onto just 2 grids. This ultimately leads to the qualitative improvement in computation time of the FPR algorithm.

\subsection{Main contributions}

Note that this paper is not about path-planning {\it per se}. It claims no new contribution whatsoever to the extensive science of path-planning~\cite{Latombe12}. Its novel contribution is entirely directed at the efficient computation of the risk of collision.

Our principal contributions are as follows.
\begin{enumerate}
\item A linearisation of the Laugier integrals scheme gives a close approximation and a bound on the risk, and allows the entire computation to be done over just two grids, regardless of the number of obstacles. That reduces computational complexity from $O(NK)$ to $O(N+K)$, for $K$ point obstacles and $N$ paths. So far, this applies only to point obstacles, not obstacles of finite size.
\item The Laugier integrals can however be extended by means of Minkowski sums to apply to obstacles of finite size.
That,  together with a new ``convolution trick'', leads to the FPR algorithm for computing a bound on the risk of collision with finite, tethered obstacles. The computational complexity of the FPR algorithm is $O(N+K)$, as desired.
\item Simulations quantify the difference between the FPR bound on risk and the true risk, under various circumstances.
\item Simulations with simulated and real data show that the $O(N+K)$ computational complexity does indeed lead to substantial reductions in practical computation times.
\end{enumerate}

\section{EFFICIENT COMPUTATION OF BOUNDS ON COLLISION PROBABILITIES}

Given the shapes and uncertain location of obstacles in an environment, the problem is to estimate the risk of collision for a set of candidate paths. This risk computation uses the probability distributions for encroachment by obstacles on the path swept by a moving robot, during a given time interval. Our starting point is the work by Laugier and collaborators~\cite{r10b,r10} who propose a probabilistic framework of this sort. This section extends the framework and develops an efficient algorithm for computing risk.

\subsection{Probabilistic Obstacle Framework}
Consider an environment consisting of a set of $K$ point obstacles. The obstacles are typically detected by perception modules whose outputs are uncertain (by design), so the position of each obstacle is a random variable given by the density function $p_k({\bf r})$ where ${\bf r}=(x,y)\in\Re^2$. 
Then the probability of collision between the robot and the $k^{\rm th}$ point obstacle can be written~\cite{r10} as
\begin{equation}
  \label{e:Laugier0}
      P_{\rm D}(k) = \int_A p_k({\bf r})
  \end{equation}
where $A$ is the swept area of the robot along a path $\pi$ and over a time interval $t \in [0,T]$.
Now the total probability of collision $P_{\rm D}$ is computed~\cite{r10} as
\begin{equation}
  \label{e:PD}
      P_{\rm D} = 1 - \prod_{k=1}^K (1-P_{\rm D}(k)),
  \end{equation}
which must be recomputed for {\textit{each}} swept path $A$ of $N$ candidate paths. However, we propose instead a bound $P_{\rm D}\leq \bar{P}_{\rm D}$ that can be computed as  
\begin{equation}
      \bar{P}_{\rm D} = \sum_{k=1}^K P_{\rm D}(k) .
      \label{e:PDbar}
\end{equation}
Moreover, when $P_{\rm D}\ll 1$, as we expect in practical, relatively safe situations, the bound $\bar{P}_{\rm D}$ is tight.
The bound can be computed  efficiently, exploiting the linearity of (\ref{e:PDbar}) cf. (\ref{e:PD}), to calculate
  \begin{equation}
	  \bar{P}_{\rm D} =  \int_A G({\bf r}) \mbox{   where   } G = \sum_{k=1}^K p_k ({\bf r}) .
          \label{e:FDG}
  \end{equation}
The computation is then trivially $O(N+K)$ not $O(NK)$, since $G$ in (\ref{e:FDG}) can be precomputed and re-used for all candidate paths $A$. However, with obstacles of finite size (as opposed to point obstacles) achieving  $O(N+K)$ complexity is no longer trivial, as we see next.  

\subsection{Finite obstacles}
\begin{figure}[htbp]
\centering
\vspace{3mm}\includegraphics[width=0.45\textwidth]{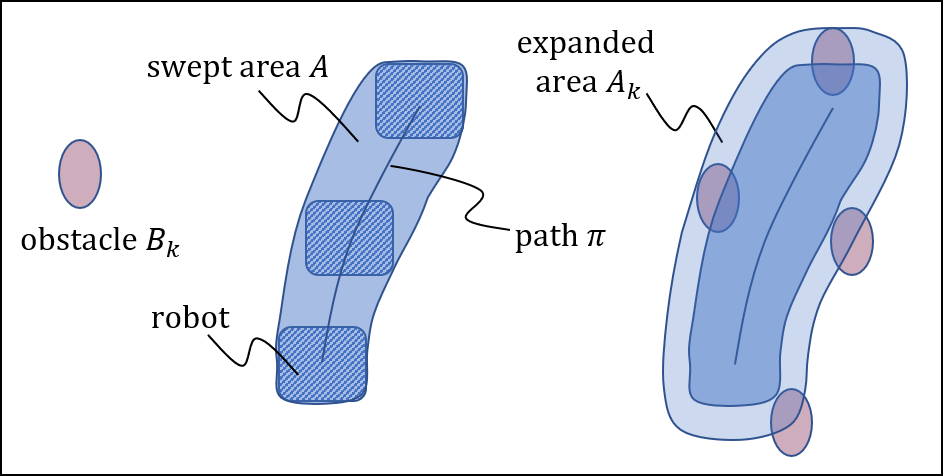}
\caption{\small {\bf Minkowski Sum.} Path $\pi$ with swept area $A$, dilated by Minkowski sum with obstacle $B_k$, to give the expanded swept area $A_k$.}\label{fig2}
\end{figure}

Generally for obstacles that are not just points but have finite area, (\ref{e:Laugier0}) has been extended~\cite{r10b}  for the case of circular obstacles by ``adding on'' the obstacle radius to the robot $A$.
More generally, for an obstacle shape $B_k \subset \Re^2$, situated at the origin, Minkowski sum can be used to expand the robot shape $A$.   At a general position ${\bf r}$, the displaced obstacle is
\begin{equation}
  B_k({\bf r}) = B_k + {\bf r} = \{ {\bf r} + {\bf r}' : {\bf r}' \in B_k \}.
\end{equation}
So the probability of collision with the obstacle can be rewritten as
  \begin{equation}
      P_{\rm D}(k) = \int_{A_k} p_k({\bf r}) ,
      \label{Laugier_int}
  \end{equation}
  where $A_k = A \oplus B_k$, the Minkowski sum of the robot shape and the obstacle shape, as in figure~\ref{fig2}. We term this equation (\ref{Laugier_int}) the {\em Laugier Integral}.
  
In a search-based algorithm, the Minkowski sums for $A_k$ must then be recomputed for each of $N$ candidate swept paths $A$, and for every obstacle $B_k$. We would therefore like to find a way to replace this naive $O(NK)$ computation, by an efficient $O(N+K)$ computation --- as was done above for point obstacles, but now in the finite obstacle case.
\subsection{Minkowski Sum and the ``Convolution Trick''}
\begin{figure}[htbp]
\centering
\vspace{3mm}\includegraphics[width=0.45\textwidth]{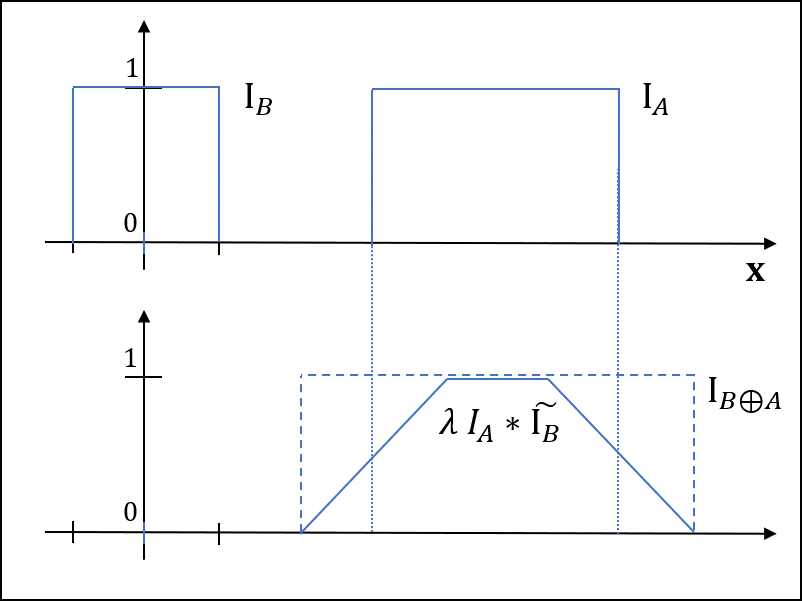}
\caption{\small {\bf Convolution and the Minkowski sum.} Illustration in 1D of the fact that the indicator of a Minkowski sum is not generally equal to the convolution of the two indicators, but they do share the same support.}\label{fig3}
\end{figure}

Note that the integral in (\ref{Laugier_int}) can be rewritten as the mathematical convolution of two functions, evaluated at the origin:
\begin{equation}
  \label{e:LaugierConvol}
      P_{\rm D}(k) = [{\rm I}_{A_k} \ast \tilde{p}_k({\bf r})]({\bf 0}) ,
  \end{equation}
where ${\rm I}_S$ denotes the indicator function of the set $S$, $\tilde{f}$ denotes the reflection of a function, i.e., $\tilde{f}({\bf r}) = f(-{\bf r})$, and the notation $[\ldots]({\bf r})$ means that the function defined in square brackets is  evaluated at the location ${\bf r}$. (So in this instance (\ref{e:LaugierConvol}), the function in square brackets is a convolution of two functions, which is then evaluated at the origin ${\bf r}= {\bf 0}$.)
There is also a known connection~\cite{ab1} (see also~\cite{r20b, r20}) between  the convolution of the indicator functions of two sets, and the Minkowski sum of the two sets, as follows:
  \begin{equation}
       X \oplus Y = \mbox{supp}({\rm I}_X \ast \tilde{\rm I}_Y) ,
  \end{equation}
where $\mbox{supp}(f)$ is the support of the function $f$. In particular, 
  \begin{equation}
      A_k = \mbox{supp}({\rm I}_A \ast \tilde{\rm I}_{B_k}) .
  \end{equation}
It is not generally the case that the indicator of a Minkowski sum is simply equal to the (normalised) convolution of the two indicator functions (see figure~\ref{fig3}). Nonetheless, over a restricted portion of the domain, corresponding to the case when the obstacle $B_k$ lies {\textit{inside}} the robot path $A$, equality does hold:
  \begin{equation}
      {\rm I}_{A_k}({\bf r}) = \lambda_k ~ [{\rm I}_A \ast \tilde{\rm I}_{B_k}]({\bf r}) ~\mbox{when}~ B_k({\bf r}) \subset A ,
      \label{conv_eq_1}
  \end{equation}
where $\lambda_k = \frac{1}{area(B_k)}$. The expression on the right of this equation is everywhere positive as it is a convolution of indicator functions which are positive.
This gives us a formula for ${\rm I}_{A_k}$ when $B_k({\bf r}) \subset A$. Next, we need the corresponding formula for the complementary case when $B_k({\bf r}) \not\subset A$.

\subsection{Contour convolution}
For the complementary component of the Minkowski sum, where $B_k({\bf r}) \not\subset A$, then ${\rm I}_{A_k}$ can be bounded using a convolution of the bounding contours of obstacles $\partial{B_k}$, and of the robot $\partial{A_k}$. This leads to an upper bound on any integral of the form $\int_{A_k} f({\bf r})$, and in particular on the collision probability (\ref{Laugier_int}).

\begin{figure}[htbp]
\centering
\vspace{3mm}\includegraphics[width=0.45\textwidth]{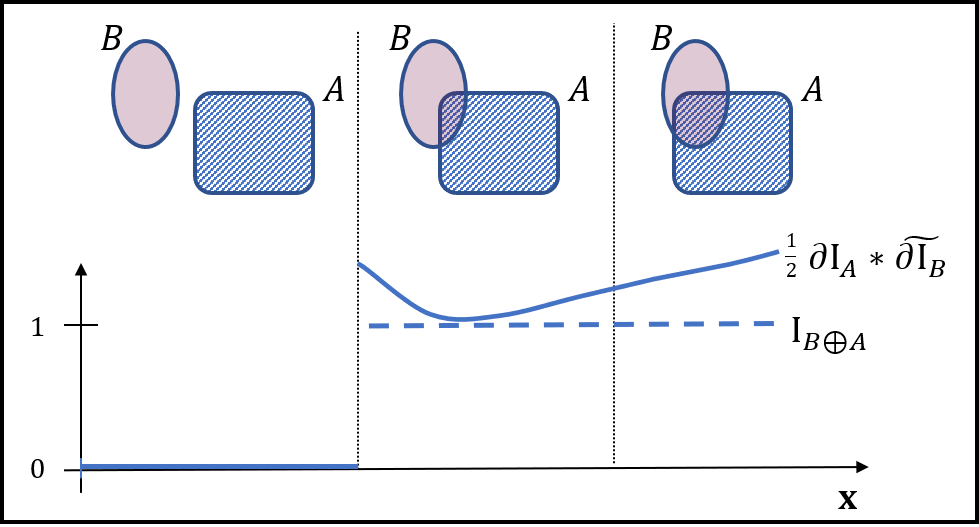}
\caption{\small {\bf Contour convolution.} Approximating the indicator function of the Minkowski sum of sets $A$ and $B$ via contour convolution.}
\label{fig4}\label{f:contour-convolution}
\end{figure}

Given the set $A$, we define the {\em delta function ridge} around its boundary $\partial{A}$ as:
  \begin{equation}
      \partial{A_\sigma}({\bf r}) = |\nabla g_\sigma({\bf r}) \ast {\rm I}_A ({\bf r})|
  \end{equation}
  where $g_\sigma({\bf r})$ is a normalised, isotropic, $2$D Gaussian function with a (small) diameter $\sigma$.
 Similarly, we define $\partial{B}_{k,\sigma}({\bf r})$ as the delta function ridge around $\partial{B}_{k}$.
Now, we claim that the indicator function for the Minkowski sum is bounded in the complementary condition, and in the limit that $\sigma \rightarrow 0$, by the convolution of these two delta function ridge functions, as follows:
  \begin{equation}
  	{\rm I}_{A_k}({\bf r}) \le \frac{1}{2} [\partial{A_\sigma} \ast \tilde{\partial{B}}_{k, \sigma}]({\bf r}) ~\mbox{when}~ B_k({\bf r}) \not\subset A    
    \label{conv_eq_2}
  \end{equation}
This is illustrated in figure~\ref{fig4}, and proved later.
\subsection{Combined bound for the FPR algorithm}

As with (\ref{conv_eq_1}), the right hand side of the inequality (\ref{conv_eq_2}) is everywhere positive. So, now the mutually complementary expressions (\ref{conv_eq_1}) and (\ref{conv_eq_2}) can be combined into a single bound on the indicator function of the Minkowski sum:
  \begin{equation}
  	{\rm I}_{A_k}({\bf r}) \le \frac{1}{2} \left[\partial{A_\sigma} \ast \tilde{\partial{B}}_{k, \sigma} \right]({\bf r}) + \lambda_k ~\left[{\rm I}_A \ast \tilde{\rm I}_{B_k}\right]({\bf r}) .
    \label{conv_eq_3}
  \end{equation}
As this bound holds everywhere, we can simply write
  \begin{equation}
  	{\rm I}_{A_k} \le \frac{1}{2}  \partial{A_\sigma} \ast \tilde{\partial{B}}_{k, \sigma} + \lambda_k ~{\rm I}_A \ast \tilde{\rm I}_{B_k} .
    \label{conv_eq_4}
  \end{equation}
Returning to the earlier expression~(\ref{e:LaugierConvol}) for the collision probability, we have
  \begin{equation}
  	P_{\rm D}(k) \le \left[ \left( \frac{1}{2} \partial{A_\sigma} \ast \tilde{\partial{B}}_{k, \sigma} + \lambda_k {\rm I}_A \ast \tilde{\rm I}_{B_k}\right) \ast \tilde{p_k} \right] ({\bf 0}) .
    \label{conv_eq_5}
  \end{equation}
Now using the associativity of the convolution operator, this can be rewritten as
  \begin{equation}
  	P_{\rm D}(k) \le   \frac{1}{2} \left[ \partial{A_\sigma} \ast \tilde{\partial{B}}_{k, \sigma} \ast \tilde{p_k} \right] ({\bf{0}}) + \lambda_k \left[ {\rm I}_A \ast \tilde{\rm I}_{B_k} \ast \tilde{p_k}\right]({\bf 0}) .
    \label{conv_eq_6}
  \end{equation}
This can equivalently be written as
  \begin{equation}
  	P_{\rm D}(k) \le \int \partial{A_\sigma} \frac{1}{2} \left(\partial{B_{k,\sigma}} \ast {p_k}\right) + \int  {\rm I}_A \left( \lambda_k {I}_{B_k} \ast {p_k} \right) .
    \label{bound_eq_1}
  \end{equation}
Finally, summing up over obstacles as in (\ref{e:PDbar}), the bound $F_{\rm D}$ on the number of collisions is given by:
  \begin{equation}
  	\bar{P}_{\rm D} \leq F_{\rm D} = \int \partial{A_\sigma}({\bf r}) \partial G_\sigma({\bf r}) + \int {\rm I}_A({\bf r}) G({\bf r})
    \label{bound_eq_2}
  \end{equation}
where $\partial G_\sigma$ and $G$ are:
	\begin{align}
    \label{Gsigma_eq}	\partial G_\sigma &= \frac{1}{2} \sum_k \partial{B_{k,\sigma}} \ast p_k \\
       \label{G_eq} G &= \sum_k \lambda_k {\rm I}_{B_k} \ast p_k .
    \end{align}
Note that $G$ and $\partial G_\sigma$ are independent of $A$ and {\textit{do not}} need to be {\em recomputed} every time $A$ changes. So the repeated computation of the bound (\ref{bound_eq_2}), for $N$ different swept paths $A$, would indeed have complexity $O(N+K)$.
This combined  ``convolution trick'' gives the FPR method to calculate the bound on the number of collisions which is summarised in Algorithm~1.

The two equations (\ref{Gsigma_eq}) and (\ref{G_eq}) combine the full set of obstacles (together with their location distributions) onto 2 grids or planes. This non-trivial combination of $K$ obstacle shapes and distributions onto just 2 grids is what gives the FPR algorithm its increased efficiency. However, it is important to note that this is {\em not simply a union of obstacles}. It is a  complex combination of shapes, outlines and location distributions, which is by no means obvious, but is derived and justified in this paper by means of the convolution trick.

\subsection{Proof of the Contour Convolution Formula}
We show that the indicator function for the Minkowski sum is indeed bounded, in the limit $\sigma \rightarrow 0$, as in inequality (\ref{conv_eq_2}).
For values of ${\bf r}$ such that $B_k({\bf r}) \cap A = \phi$, both sides of the inequality in (\ref{conv_eq_2}) are $0$, in the limit. Elsewhere $B_k({\bf r}) \not\subset A$, so the contours $\partial A$ and $\partial B_k$ must intersect at least twice. In that case, the convolution
\begin{equation}
	[\partial{A}_\sigma \ast \tilde{\partial{B}}_{k,\sigma}]({\bf r}) = 
    	\int_{{\bf r}'} \partial{A_\sigma}({\bf r}') \partial{B}_{k,\sigma} ({\bf r}' -{\bf r})
    \label{proof_eq_2}
\end{equation}
integrates across two or more contour intersections. The integral at each intersection of two smooth contours (crossing at an angle $\theta$) has the general form
\begin{equation}
	J = \int\int g_\sigma(x) g_\sigma(x \cos\theta + y \sin\theta) \,{\rm d}x \,{\rm d}y .
    \label{proof_eq_3}
\end{equation}
Now as $g_\sigma$ is a normalised Gaussian, $\int_x g_\sigma(x)\,{\rm d}x = 1$, and applying this above  in $y$ (with a straightforward substitution), and in $x$, yields
\begin{equation}
	J = \frac{1}{\sin\theta} \ge 1 ,
    \label{proof_eq_4}
\end{equation}
%
so the integral (\ref{proof_eq_2}) accumulates a value of at least 1 for each contour intersection. Therefore, with 2 or more intersections, the right hand side of inequality~(\ref{conv_eq_2}) is at least $\frac{1}{2}\times 2=1$, compared with the value of 1 for the indicator function ${\rm I}_{A_k}({\bf r})$ on the left hand side, so the inequality does indeed hold.
\RestyleAlgo{ruled}
\begin{algorithm}
  \KwData{$N$~instances~of~path~$A$, $B_{1:K}$, $p_{1:K}({\bf r})$, $\sigma$}
  \KwResult{$F_{\rm D}$}\;
  \textit{// Compute $\partial G_\sigma$}\;
  \For{k in 1 to K} {
    $\partial{B_{k, \sigma}}({\bf r}) = |\nabla g_\sigma({\bf r}) \ast {\rm I}_{B_k} ({\bf r})|$
  }
  \;
  $\partial G_\sigma({\bf r}) = \frac{1}{2}\sum_{k=1}^{K}{\partial{B_{k, \sigma}}({\bf r}) \ast p_{k}({\bf r})}$\;
  \;
  \textit{// Compute $G$}\;
  $G({\bf r}) = \sum_{k=1}^{K}{\frac{1}{\text{area}(B_k)}{\rm I}_{B_k}({\bf r}) \ast p_{k}({\bf r})}$\;
  \;
  \For{{\rm each}~A} {
    \textit{// Compute $\partial A_\sigma$}\;
    $\partial{A_\sigma}({\bf r}) = |\nabla g_\sigma({\bf r}) \ast {\rm I}_A ({\bf r})|$\;
    \;
    \textit{// Integrate over a box around A with margins $4\sigma$}\;
    $F_{\rm D} = \int_{\mathbb{R}^2}{\partial{A_\sigma}({\bf r})\partial G_\sigma({\bf r}) + {\rm I}_A({\bf r})G({\bf r})}$
  }
  \;
  \caption{\small {\bf FPR algorithm} for the bound on   collision risk  given $N$ paths, and $K$ obstacles.}
\end{algorithm}

\section{Results}

In this section we demonstrate that randomly generated trajectories of an SE(2) robot can be efficiently labelled by the FPR algorithm, according to the bound $F_{\rm D}$ on the risk of collision for each path. The FPR algorithm is agnostic as to the motion planner used to synthesise the candidate trajectories and is generally compatible with state of the art methods for motion planning \cite{katrakazas2015real}. We use an off-the-shelf Closed Loop variant CL-RRT~\cite{kuwata2009real} of the RRT algorithm~\cite{lavalle1998rapidly}
to generate candidate paths in the environment, drawn from the kinodynamic model for a particular robot. This has the advantage of generating typically smooth paths, that are plausible paths for that robot. Then the risk bound is calculated for each generated path.

First our results demonstrate the FPR algorithm for a simulated environment, then for a real environment taken from an aerial view, and finally a substantial dataset of 7481 birds-eye views each with several goals and multiple trajectories for each goal --- 240,498 trajectories in all. In each case, higher risk paths take tighter lines around obstacles, as would be expected. We show: i) how close the bound on risk is to the true risk; and ii) that  the use in FPR of the convolution trick, which improves computational complexity from $O(NK)$ to $O(N+K)$ as explained earlier, leads to substantial reductions in practical computation times.

\subsection{Simulated Environments}
We first use a $2$D simulation in which a rectangular SE(2) robot of size $2{\rm m} \times 4{\rm m}$ navigates along continuous paths, defined as the progression of $(x,y,\theta)$ pose over time (though our visualisations only depict the centroid). A simulation scenario is defined as a collection of obstacles within the environment, each specified as a shape (a subset of $\Re^2$), and pose, together with positional uncertainty, as well as start and goal poses for the ego vehicle.

\begin{figure}[htbp]
\centering
\vspace{3mm}\includegraphics[width=0.45\textwidth]{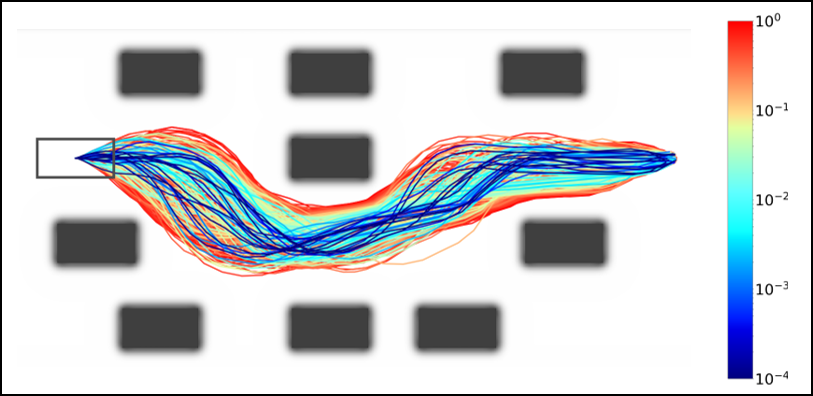}
\caption{\small {\bf Visualisation of paths.} The figure shows paths for a simulated environment. A set of candidate paths are generated with fixed start and end poses. The dark objects are obstacles whose position is known with an uncertainty visualised here as a shaded halo. Risk bounds $F_{\rm D}$ on each path are represented by different colours, on the logarithmic scale shown. Safer paths maintain greater clearance around obstacles as expected. Some risky paths (bottom) involve squeezing through a narrow gap.  }\label{f:sim-results}
\end{figure}

The simulated scenario shown here in figure~\ref{f:sim-results} resembles sections of a car park; 400 paths are generated at random by CL-RRT. The uncertainty over each obstacle's position is modelled as a two-dimensional Gaussian distribution with standard deviation $0.3$m, which is 15\% of each obstacle's width.
In figure~\ref{f:sim-results}, paths with lower $F_{\rm D}$ are seen to maintain a greater clearance from the  obstacles, as expected. 

In figure~\ref{f:trajectory-risk} an aerial view of a car park is shown, with a set of candidate paths generated by CL-RRT, between fixed start and end points. Obstacle vehicle shapes are represented as bounding rectangles. Error in estimated position of the obstacle-cars is Gaussian with standard deviation of 0.3m.
The candidate paths are coloured according to the computed value of the bound on collision risk. This turns out to include safer paths with collision risk down to $10^{-5}$ and below, and riskier paths, above $10^{-2}$ risk of collision, that involve squeezing through a narrow gap.

\begin{figure}[htbp]
\centering
\vspace{3mm}\includegraphics[width=0.48\textwidth]{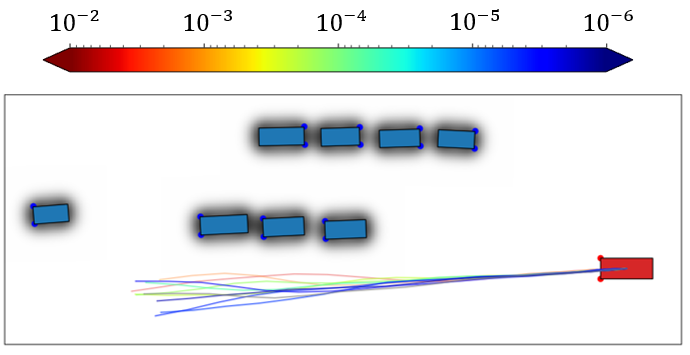}
\caption{\small {\bf KITTI birdseye data.} Example  view from  KITTI dataset of 7,481 birdseye views of traffic scenes. Obstacles shown in blue, each have simulated Gaussian uncertainty with standard deviation 0.7m. A goal is chosen automatically. Possible paths for an SE(2) robot (red) are shown. FPR bound on collision risk is displayed on the colour scale shown. 
}\label{f:KittiTrajectories} 
\end{figure}

Finally, for the sole purpose of researching the behaviour of the FPR algorithm, a larger dataset derived from the birds-eye view KITTI collection~\cite{KITTI13} is used. Each scene contains a number of vehicles, obstacles $B_k$ that are represented as rectangles, with positions labelled and assumed here to have Gaussian error with standard deviation of 0.7m. One example view, from the total of 7481 birdseye views, is illustrated in figure~\ref{f:KittiTrajectories}.
In each scene, an SE(2) robot is given an initial position, and several goals are chosen, automatically. Then up to 10 paths per goal are generated --- a total of 240,498 paths. Shapes of obstacles are assumed known, and in practice this could be achieved by recognition of known objects such as vehicles.

\subsection{Performance Evaluation}
\begin{figure}[htbp]
\centering
\vspace{3mm}\includegraphics[width=0.45\textwidth]{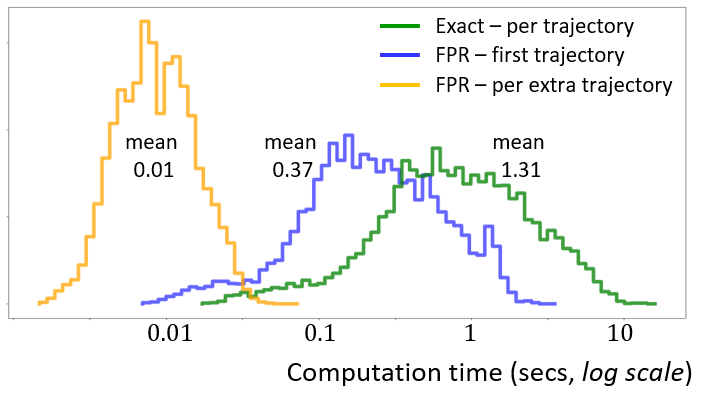}
\caption{\small {\bf Performance of FPR bound computation} over the KITTI birdseye data, compared with exact (\ref{Laugier_int}) computation of risk.  
}\label{f:timings} 
\end{figure}

Simulation results given here illustrate the computational benefits of the FPR approach for evaluating bounds on risk. In figure~\ref{f:timings}, we present empirical data regarding the computational efficiency of our method compared to the exact computation of the integral in (\ref{Laugier_int}). Computing the FPR bound is, on average, significantly faster than exact computation. Even for the first path, FPR is more than 3 times as efficient on average, thanks to the use of efficiently implemented convolution, in place of Minkowsi sum. For subsequent paths FPR is on average two orders of magnitude more efficient, at 10ms per path. This is consistent with $O(N+K)$ complexity c.f $O(NK)$ complexity (for $N$ evaluated paths), as expected from theory.

\subsection{How tight is the bound on collision risk?}

\begin{figure}[htbp]
\centering
\vspace{3mm}\includegraphics[width=0.45\textwidth]{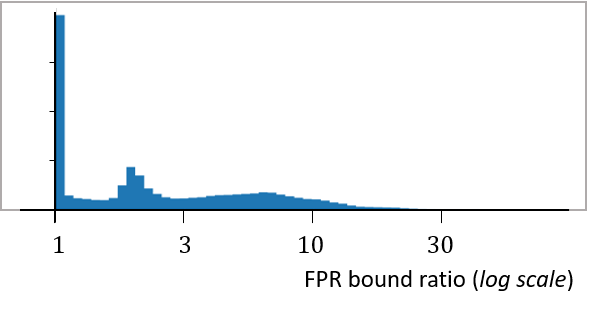}
\caption{\small {\bf Tightness of bounds -- KITTI data.} Bounds $F_{\rm D}$ on collision risk are computed for all 240,498 trajectories and compared with exact risk. The risk bounds have an average ratio of 2.72 times the exact risk, and that ratio is distributed as shown in the histogram.}
\label{f:KittiSlack} 
\end{figure}

It is reasonable to ask how close the FPR bound is, in practice, to the exact risk  $P_{\rm D}$. The ratio of the bound to the exact risk is evaluated over all the 240,498 paths derived from the KITTI data. The ratio is about 2.7 on average, with a distribution largely  between 1 and 10 (93\% of examples), as in figure~\ref{f:KittiSlack}.
The effect of this ratio is that the bound will lead to conservative decisions. For example if the ratio is $3$, then to achieve a desired collision risk of say $10^{-3}$ or better, setting the FPR risk to $10^{-3}$ will actually achieve a lower risk of $\frac{1}{3} 10^{-3}$. As a result, selected paths may leave more clearance from obstacles than strictly necessary. Occasionally, it is possible that no path in a certain set may have an FPR bound within the acceptable level of risk, even when a path with an acceptable level of true risk does actually exist.

\subsection{Implementation details}
We use an augmented version of the CL-RRT planner, with probabilistic sampling, similar to other approaches for heuristically biasing RRT growth~\cite{urmson2003approaches}, and choosing tree nodes for expansion according to their scores. It discretises steering and accelerator input ranges when expanding the random tree, to generate realisable trajectories, and in order to restrict abrupt steering or velocity changes. Nodes in the RRT are scored based on their proximity to the goal, and similarity to its orientation and velocity. We treat each
tethered obstacle as deterministic, just for the purposes of CL-RRT, taking the shape $B_k$ at the mean location over $p_k$.

For all simulations, space is discretized on a grid with a  resolution of 5 cm/px. All convolutions
in our implementation use a Gaussian or gradient of Gaussian kernel and so we exploit the separability property in order to perform convolutions efficiently. Additionally, we approximate the final integration step in Alg. 1 with a computationally efficient Riemann sum over the discretised
grid. (If it were desired to use non-Gaussian $p_k({\bf r})$ then the convolutions with $p_k({\bf r})$ could be done by FFT or morphologically~\cite{Samaniego19}.)

The constant for Gaussian convolution,  $\sigma=2$ grid squares, is just big enough for good numerical behaviour.
All of the numerical computations are implemented using the GPU-enabled Python module for linear algebra CuPy~\cite{cupy_learningsys2017}. The goal and trajectory data, used with KITTI data in simulations, will be made available on the web.

\section{CONCLUSIONS}

Our FPR algorithm bounds the risk of collision for candidate paths in a given environment.
It builds on a probabilistic framework for calculating collision risk, using the convolution trick to render these computations in linear time in $N$ and $K$. Amongst trajectories deemed safe enough, there would then be freedom to optimise for other criteria such as rider comfort and travel time.

Other sources of uncertainty would of course also need to be taken into account in an end-to-end implementation, such as missed detections of obstacles. The current state of the art~\cite{r23} suggests risk of the order of $10^{-3}$ for missed detections. It is to be hoped however that this improves with future advances in temporal and cross-modal fusion.

The effect of computing risk in $xy$-space, as opposed to the full $xyt$-space, is further to approximate (in fact bound) the computed risk. The bounding effect preserves safety, but in some circumstances is overly conservative and may overestimate risk, which could lead to `frozen robots'~\cite{r9}.
Future work looks at extending FPR from tethered obstacles to fully dynamic obstacles whose position evolves stochastically. Then $p_k({\bf r})$ in (\ref{Laugier_int}) would be extended to a spatio-temporal (stochastic) process as in~\cite{ab2,ab3,ab5}. Risk computation would be in the full $xyt$-space. The question is, to what extent could full Monte-Carlo computation of risk be avoided, and linear time-complexity in $N$ and $K$ be retained?


\addtolength{\textheight}{-13cm}   









\end{document}